\documentclass{article}
\usepackage{spconf, amsmath, graphicx, tabularx, multirow, stfloats, booktabs, latexsym}
\usepackage[colorlinks, linkcolor=blue]{hyperref}

\title{Self adaptive global-local feature enhancement for Radiology Report Generation}

\name{Yuhao Wang$^{1}$, Kai Wang$^{2}$, Xiaohong Liu$^{3}$, Tianrun Gao$^{1}$, Jingyue Zhang$^{1}$, Guangyu Wang$^{1, \star}$\thanks{$^{\star}$Corresponding author:guangyu.wang24@gmail.com }}

\address{$^{1}$State Key Laboratory of Networking and Switching Technology,\\
Beijing University of Posts and Telecommunications, Beijing 100876, China\\ 
$^{2}$Peking-Tsinghua Center for Life Sciences, Peking University, Beijing 100871, China.\\
$^{3}$Department of Computer Science and Technology, Tsinghua University, Beijing 100084, China}

\begin{document}
%
\maketitle
\begin{abstract}

Automated radiology report generation aims at automatically generating a detailed description of medical images, which can greatly alleviate the workload of radiologists and provide better medical services to remote areas. Most existing works pay attention to the holistic impression of medical images, failing to utilize important anatomy information. However, in actual clinical practice, radiologists usually locate important anatomical structures, and then look for signs of abnormalities in certain structures and reason the underlying disease. In this paper, we propose a novel framework AGFNet to dynamically fuse the global and anatomy region feature to generate multi-grained radiology report. Firstly, we extract important anatomy region features and global features of input Chest X-ray (CXR). Then, with the region features and the global features as input, our proposed self-adaptive fusion gate module could dynamically fuse multi-granularity information. Finally, the captioning generator generates the radiology reports through multi-granularity features. Experiment results illustrate that our model achieved the state-of-the-art performance on two benchmark datasets including the IU X-Ray and MIMIC-CXR. Further analyses also prove that our model is able to leverage the multi-grained information from radiology images and texts so as to help generate more accurate reports.
 
\end{abstract}
\section{Introduction}
\label{sec:intro}
Radiology reports consist of abundant abnormal observations about the medical images  are important to the clinical diagnosis. In practice, writing a comprehensive and accurate radiology report is very time-consuming and difficult \cite{jing2017automatic}. Fully automated report generation can effectively assist physicians in writing imaging reports, improving the accuracy of disease detection while reducing their workload. 

Recently, a lot of substantial progress has been made towards research on automated radiology report generation models. Most frameworks adopt the encoder-decoder architecture \cite{jing2017automatic,R2Gen,aligntransformer}, e.g., a VIT \cite{vit} image encoder and then a report decoder based on transformer decoder \cite{attention}. Nevertheless, physicians typically write radiology
reports through multi-granularity information, instead of only focused on holistic impression.
For instance, some sentences "There is minimal patchy atelectasis or early infiltrate in left upper lung zone" reflected some statements have a strong correlation with the anatomies(region features). However, some sentences like "No pneumothorax, or pleural effusion.", describing CXR from a global perspective is more likely to be inspired by global features.\par 
To explore the intrinsic relationship between different radiology reports, anatomies and holistic impression,  we propose a new Anatomy-enhanced framework (AGFNet), which mimics the human visual system and thinking process. Inspired by AnaXNet \cite{aanaxnet}, our proposed AGFNet consists of three modules including a feature extractor, an adaptive fusion module, and captioning generator.
For the feature extractor module, it consists of region and bilateral global(frontal and lateral) feature extractor. Then, an adaptive fusion of weight-based merging mechanisms is introduced to dynamically fuse multi-granularity feature which match the pathological changes degree of the patient. Finally, the captioning generator fed with multi-granularity features generates the radiology report. In summary, our approach creates disease-oriented visual features that can represent abnormal regions of the input image. Meanwhile, during model training, the self-adaptive gate can effectively learn the multi-grain relationships between images of different pathological changes and corresponding radiology reports.  
The main contributions of this paper are summarized as follows:
\begin{itemize}
\setlength{\topsep}{0pt}
\setlength{\itemsep}{0pt}
\setlength{\parsep}{0pt}
\item We propose a novel radiology report generation framework that integrates both global and local anatomical visual features
\item We develop a novel self-adaptive fusion gate module to dynamically fuse the global and region features.
\item In-depth experiment and analysis demonstrate that our proposed AGFNet model outperforms previous baselines and state-of-the-art models.
\end{itemize}

\section{Related works}
\label{sec:format}
Radiology report generation is most similar to image captioning. Some studies use Reinforcement learning obtained a good results \cite{2022reinforced, REF2021}. Image retrieval and knowledge embedding also achieved great success in this domain\cite{KnowledgeMR, 2022knowledge}.
The encoder-decoder architecture, generally used in image captioning, is the most successful architecture in radiology report generation\cite{jing2017automatic, R2Gen, cmn, aligntransformer}.
Most methods based on encoder-decoder leverage the architecture like vision transformer\cite{vit} as the image encoder\cite{R2Gen, cmn, aligntransformer}, which split an input image as a series of image patches. Then, decoders\cite{LSTM,attention}generate the target sequence based on the all patch features. All these methods lead to the serious destruction of region information, which result in the model failed to learn the characteristics of region pathological changes and utilize the strong relationship between different anatomies.

\section{Methods}
\label{sec:pagestyle}
\subsection{Model overview}

\begin{figure*}[hb]
\centering

\vspace{-0cm}  
\setlength{\abovecaptionskip}{-10cm}   
\setlength{\belowcaptionskip}{-50cm}   
\includegraphics[width=0.76\textwidth, height=0.3\textwidth]{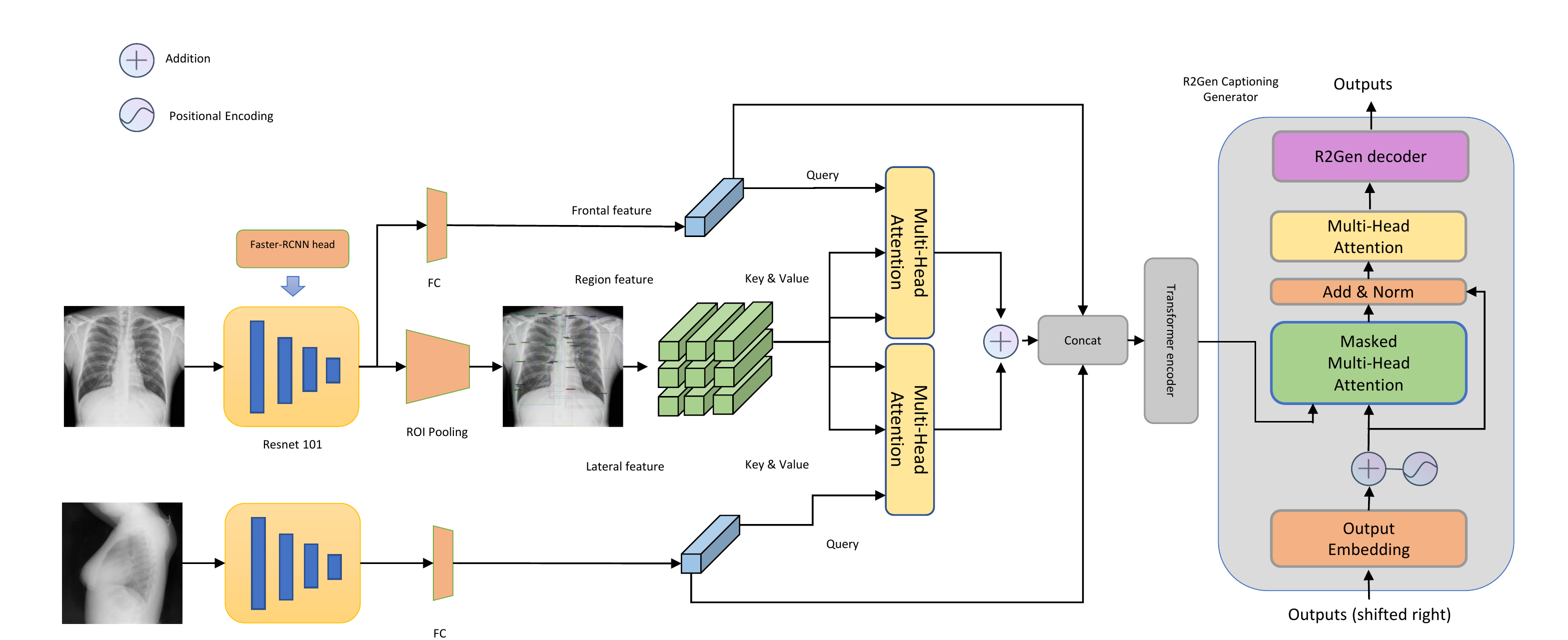}
\caption{Illustration of the AFGNet, the feature extractor acquires local and global features of the CXR. Then, the self-adaptive fusion gate generates multi-granularity features related to the degree of pathological changes, and finally generates diverse radiological reports through the captioning generator.}
\label{Model Overview}
\end{figure*}

Our model consists of three components: The global and region feature extractor, the self-adaptive fusion gate can dynamically fuse global and regional features according to the severity of the pathological changes, and the captioning generator. Our proposed method is shown Fig. \ref{Model Overview}.

\subsection{Global and Region feature extractor}
The goal of the regional feature extractor is to accurately locate the most prominent anatomy structures and obtain their robust features differentiated by pathological changes. These features have fine-grained differences owing to the severity level of the pathological changes. Consequently, we pretrained ResNet101 Backbone Faster-RCNN \cite{Faster-RCNN} to locate the anatomies and apply the RoIPooling \cite{Faster-RCNN} to extract anatomy regions of interest (ROIs) which frequently appear in radiology reports, i.e., cardiac silhouette, left lower lung, left mid lung zone, trachea,  mediastinum, etc are extracted, denoted as$$V_{region}=\left\{v_{cardiac}, v_{trachea}, v_{mediastinum}, ...\right\} \in {\mathbf{R}^{N\times d}} $$where N is the number of anatomies, each with d dimensionality.
For global features, the global feature of frontal view is generated by the backbone of the region feature generator, the following is an fully connected layers layer to flatten the feature map. The feature of lateral view is acquired by the other ResNet101 backbone. Such a design is to reduce the computational overhead. Briefly, the global feature is formulated as
$$V_{global}=\left\{V_{frontal}, V_{lateral}\right\}\in {\mathbf{R}^{2\times d}}$$
As shown in Table \ref{table:1}, the region feature extractor demonstrate an outstanding result.
We define right lung (RL),  right apical zone (RAZ),  right upper lung zone (RULZ),  right mid lung zone (RMLZ),  right lower lung zone (RLLZ),  right costophrenic angle (RCA),  left lung (LL),  left apical zone (LAZ),  left upper lung zone (LULZ),  left mid lung zone (LMLZ),  left lower lung zone (LLLZ),  left costophrenic angle (LCA),  mediastinum (Med),  upper mediastinum (UMed),  cardiac silhouette (CS) and trachea (Trach).mAP is applied to evaluate the performence of region feature detector.

\begin{table}
\centering
\caption{Detection result of region feature extractor}
\label{table:1}
\begin{tabular}{ |p{4em}|p{4em}|p{4em}|p{4em}|  }
\hline
\multicolumn{4}{|c|}{Detection Results} \\
\hline
Anatomy&mAP@0.5&Anatomy&mAP@0.5 \\
\hline
RL&0.950&RAZ&0.900\\
RAZ&0.912&RMZ&0.725\\
RLLZ&0.905&LULZ&0.950\\
LAZ&0.887&LF&0.776\\
LMLZ&0.898&LLLZ&0.852\\
Med&0.773&UMed&0.886\\
CS&0.920&TR&0.975\\
 \hline
\end{tabular}
\end{table}

\subsection{Self-Adaptive Fusion Gate}
In clinical practice, there is a consensus that the radiologist writes the radiology report in different granularity according to different degree of pathological changes. This situation can be demonstrated in the dataset such as IU-X-ray or MIMIC, e,g, most description in some normal cases is like “No pneumothorax, or pleural effusion”, which is more likely concluded from a global impression of the whole image. However, some cases with pathological changes are often described in more detail, including many fine-grained descriptions of abnormalities occurring in the associated 
anatomies, are highly correlated with features of several pathological regions. Consequently, in the radiologist decision process, the region feature and global feature dynamical aggregated based on image morphology.
Inspired by attention mechanism\cite{attention}, we propose the self-adaptive Fusion gate to dynamically fuse the multi-granularity features. The self-adaptive fusion gate adopted region features, front view features, lateral view features which can be indicated as $V_{r},V_{f},V_{l}$. We use the global vector $V_ {view}\in\mathbf{R}^{1\times d} $ which is a element of $V_{global}=\left\{V_{f}, V_{l}\right\}$ represents the pathological manifestation as the $Q$ and region feature $V_ {r} \in \mathbf{R}^{N\times d}$ as a set of key-value pairs $K,V$, aims at creating shortcuts between every region feature and the global feature. We adopted multi-head attention $MHA$ as the main structure. The output is a weighted sum of the values, where the weight assigned to each value is determined by the dot-product of the gloabl feature with all region features:
$$ 
MHA(Q, K, V)= softmax(\frac{QK^T}{\sqrt{d_k}})V
$$

For the purpose of dynamically merging the biliteral view selected region features, the weight coefficients of the two parts are considered to be added. Therefore, the fused region feature $V_{\bar{r}}$ is acquired by:
$$
V_{\bar{r}}=MHA(V_{f},V_{r},V_{r})+MHA(V_{l},V_{r},V_{r})
$$

Finally, the self-adaptive fusion gate concatenate the fused region features and global features:
$$
V_{MG}=[V_{\bar{r}};V_{f};V_{l}]
$$
The fusion gate is trained to learn how to dynamically control the fusion degree of region features, and global features from the front view and lateral view. For normal cases, global features was more reserved and the region features are given less weight as the model need to generate more descriptive sentence about the holistic health status. For a diseased case with pathological changes occur in different anatomy regions, the self-adaptive fusion gate endow higher coefficients to region features which contain visual information more relevant to important pythology. Finally, we adopted the transformer encoder \cite{attention} to learn the relationship between different anatomy regions.

\subsection{Captioning Generator}
As R2Gen\cite{R2Gen} adopted a hidden memory matrix in captioning generator to restore the abnormal pattern information demonstrates its excellent performance, we adopted the R2Gen decoder as our captioning generator, aiming at correcting and updating the abundant information according to the different lesions occurring in different anatomies.

\begin{table*}[t]
\caption{Results from the IU-Xray and MIMIC-CXR datasets. The best values are highlighted in bold.}
\centering
\scriptsize
\label{tab:main_results}
\setlength\tabcolsep{10pt}

\renewcommand{\arraystretch}{1.5} 
\begin{tabular}{clccccccc}
\toprule  
\textbf{Dataset} & \textbf{Method} & \textbf{BL-1} & \textbf{BL-2} & \textbf{BL-3} &\textbf{BL-4}  &\textbf{RG-L} & \textbf{MTOR}  \\
\midrule

\multirow{6}{*}{\textbf{IU-Xray}}&$R2Gen$~\cite{R2Gen} &0.470 &0.304 &0.219 &0.165 &0.371 &0.187 \\ 
\multirow{6}{*} &$KERP$~\cite{KERP} &0.482 &0.325 &0.226&0.162 &0.339 & -\\
\multirow{6}{*} &$CMN$~\cite{cmn} &0.473 &0.305 &0.217 &0.162 &0.378 & 0.186\\
\multirow{6}{*} &$AlignTransformer$~\cite{aligntransformer} &0.484 &0.313 &0.225&0.168 &0.379 &0.204\\
\multirow{6}{*}&$M2TR$~\cite{M2TR}&0.486 &0.317 &0.232 &0.170 &0.390 &0.192 \\
\cline{2-8}
\multirow{6}{*}  &$Ours$  &\textbf{0.505} &\textbf{0.345} &\textbf{0.243} &\textbf{0.176} &\textbf{0.396} & \textbf{0.205} \\
 \midrule

\multirow{6}{*}{\textbf{MIMIC}} &$R2Gen$~\cite{R2Gen} &0.353 &0.218 &0.145 &0.103 &0.277 &0.128\\

\multirow{6}{*}{\textbf{-CXR}} &$PPKED$~\cite{PKED} &0.360& 0.224 &0.149 &0.106 &0.284&0.149 \\
\multirow{6}{*} &$CMN$~\cite{cmn} &0.334 &0.217& 0.140 &0.097 &0.281&0.142\\
\multirow{6}{*} &$AlignTransformer$~\cite{aligntransformer} &0.378 &0.235&0.156 &0.112 &0.283&\textbf{0.158}\\
\multirow{9}{*} &$M2TR$~\cite{M2TR} &\textbf{0.378}& 0.232 &0.154 &0.107 &0.283&0.145\\
\cline{2-8}
\multirow{6}{*}  &$(Ours)$  &0.363 &\textbf{0.235} &\textbf{0.164} &\textbf{0.118} &\textbf{0.301} & 0.136\\
\bottomrule 
\end{tabular}
\end{table*}

\section{EXPERIMENTS}
\label{sec:majhead}

In this section, we describe the foundation for our research, 
including benchmark datasets, experimental settings, and evaluation metrics. Through rigorous ablation experiments, we have verified the effectiveness of each component.

\subsection{Datasets, Metrics and Implementation details}
\label{ssec:subhead}

We conduct experiments on MIMIC-CXR \cite{MIMIC} and IU-XRay \cite{IU-Xray}, which are the most widely used in radiology report generation. For a fair comparison with existing models, we use the standard evaluation toolkit \cite{cococaption} to compute widely used metrics, namely BLEU, ROUGE-L and METEOR measure how well the generated reports match the ground truth reports. We used 14 anatomy structures that frequently appear in radiology reports as the region features $V_{region} \in {\mathbf{R}^{14\times d}}$. The evaluation matrix of region feature extractor is mean of average precision when IOU threthold is 0.5 (mAP@0.5), which is commonly used to analyze the performance of object detection. The Adam optimizer is adopted and a batch size of 128. The feature dimension d=2048, with learning rate of 2e-4 training for 100 epochs. The data used to pre-train the region feature extractor is Chest Genome \cite{chestgenome} . The details of MIMIC-CXR and IU-XRay are shown in Table \ref{table:datasets}.
\label{dataset description}
\begin{table}[t]
\footnotesize
\centering
\setlength{\tabcolsep}{1mm}{\begin{tabular}{@{}l|rrr|rrr@{}}
\toprule
\multirow{2}{*}{Dataset} & \multicolumn{3}{c|}{IU X-Ray}                                                                     & \multicolumn{3}{c}{MIMIC-CXR}                                                                   \\ \cmidrule(l){2-7} 
                                  & \multicolumn{1}{c}{Train} & \multicolumn{1}{c}{Val} & \multicolumn{1}{c|}{Test} & \multicolumn{1}{c}{Train} & \multicolumn{1}{c}{Val} & \multicolumn{1}{c}{Test} \\ \midrule
Image                 & 5.2K                              & 0.7K                              & 1.5K                                & 369.0K                            & 3.0K                            & 5.2K                             \\
Report              & 2.8K                              & 0.4K                              & 0.8K                                & 222.8K                            & 1.8K                            & 3.3K                             \\
Patient               & 2.8K                              & 0.4K                              & 0.8K                                & 64.6K                             & 0.5K                              & 0.3K                               \\ \bottomrule

\end{tabular}}
\vskip -0.25em
\caption{The statistics of  training,  validation and test sets of  two benchmark datasets,  including the numbers of images,  reports and patients.}
\label{table:datasets}
\vskip -1em
\end{table} 

\subsection{Compare with Previous Work}

Based on the IU-Xray and MIMIC-CXR datasets,  we compare the our proposed AGFnet with previous methods in this section. Table \ref{tab:main_results} demonstrates that our approach significantly outperforms M2TR\cite{M2TR}, the previous best SOTA method. In IU-Xray dataset, our methods surpasses the previous sota work by 1.9\%, 3.2\%, 0.9\%, 1.3\% in BLUE-1, BLUE-2, BLUE-3, MTOR. In the MIMIC-CXR dataset, Our method obtains and achieves a greater degree of optimization than other methods in terms of BLUE-3, blue-4, and RG-L.The main results demonstrate the effectiveness of our approach and show the advantages of our approach in generating long report descriptions. The experiment shows potential of our AGFnet in learning and generating more sophisticated radiology report for cases with many pathological changes. 

\subsection{Ablation Study}
To investigate the contribution of each component to our approach, we conduct a rigorous ablation study on IU-Xray. It is obviously that the performance is greatly enhanced by two components we developed, the global and region feature extractor and the self-adaptive fusion gate. The removal of any of them will cause significant performance deterioration. As shown in Table \ref{tab:ablation_studies}, when the region feature extractor is removed, the BLUE-1 and the BLUE-2 scores decreased by \%3.5 and 4.1\% respectively, and all other assessment metrics declined with vary degrees. This proves that region feature extractor can provide richer multi-granularity visual information for captioning generator. Similarly, with the self-adaptive gate , the BLUE-1 and BLU-2 increase by 3.1\% and 2.0\%. In summary, our proposed components improve on all metrics, which demonstrate the prominent effectiveness of our approach.

\begin{table}[t]
\caption{The experimental results of ablation studies,  The best values are highlighted in bold.}
\centering
\label{tab:ablation_studies}
\setlength\tabcolsep{3pt}
\begin{tabular}{l|cccccc}
\toprule  
\textbf{IU-Xray}  & \textbf{BL-1} & \textbf{BL-2} & \textbf{BL-3} &
\textbf{BL-4}  &\textbf{RG-L} & \textbf{MTOR} \\
\midrule  

Ours  &\textbf{0.505} &\textbf{0.345} &\textbf{0.243} &\textbf{0.176} &\textbf{0.396} & \textbf{0.205} \\
w/o region  &0.470 &0.304 &0.219 &0.165 &0.371 & 0.187 \\
w/o global  &0.483 &0.311 &0.c224 &0.171 &0.387 & 0.194 \\
w/o SAGate  &0.463 &0.325 &0.240 &0.176 &0.389 & 0.202 \\

\bottomrule 
\end{tabular}

\end{table}

\section{Conclusion}
\label{sec:illust}
In our work, we propose a novel AGFNet to mine region-level semantic information by extracting regional anatomy features. At the same time, the self-adaptive fusion gate we developed can dynamically fusion the multi-granularity features according to the degree of pathological changes. Based on the MIMIC-CXR and IU-Xray datasets, experiments demonstrate that our technique not only produces reports with correct aberrant descriptions but also achieves state-of-the-art outcomes across all measures and surpasses the traditional models. The analysis further demnonstrates the effectiveness of our approach in generating detailed radiology report.
\section{Acknowledgement}
This study was funded by the National Key Research and Development Program of China (2019YFB1404804), National Natural Science Foundation of China (grants 61906105, and 62272055), Young Elite Scientists Sponsorship Program by CAST (2021QNRC001), Major Key Project of PCL (PCL2021A15).

\bibliographystyle{IEEEbib}
\bibliography{strings, refs}

\end{document}